%% file: root.tex
\documentclass[letterpaper, 10 pt, conference]{ieeeconf}  

\IEEEoverridecommandlockouts                              

\usepackage{graphics} 
\usepackage{epsfig} 
\usepackage{mathptmx} 
\usepackage{times} 
\usepackage{amsmath} 
\usepackage{amssymb}  
\usepackage{multirow}
\usepackage{colortbl}
\usepackage{booktabs}
\usepackage{makecell}
\PassOptionsToPackage{hyphens}{url}\usepackage{hyperref}
\definecolor{mygray}{gray}{.9}
\usepackage[T1]{fontenc}
\title{\LARGE \bf
Eliminating Cross-modal Conflicts in BEV Space for \\ LiDAR-Camera 3D Object Detection
}

\author{Jiahui Fu\textsuperscript{1}, Chen Gao\textsuperscript{1$\dag$}, Zitian Wang\textsuperscript{1}, Lirong Yang\textsuperscript{2}, Xiaofei Wang\textsuperscript{2}, Beipeng Mu\textsuperscript{2}, Si Liu\textsuperscript{1}
\thanks{$^{1}$Jiahui Fu, Chen Gao, Zitian Wang, and Si Liu are with Institute of Artificial Intelligence and Hangzhou Innovation Institute, Beihang University. Email: \tt{\{jiahuifu, gaochen, wangzt, liusi\}@buaa.edu.cn}}
\thanks{$^{2}$Lirong Yang, Xiaofei Wang, and Beipeng Mu are with Meituan Inc. Email: \tt{\{yanglirong, wangxiaofei19, mubeipeng\}@meituan.com}}
}

\begin{document}

\maketitle
\thispagestyle{empty}
\pagestyle{empty}

\renewcommand{\thefootnote}{\fnsymbol{footnote}}
\footnotetext[2]{Corresponding author: Chen Gao}

\input{tex/0-abstract}

\input{tex/1-intro}

\input{tex/2-relate}

\input{tex/3-method}

\input{tex/4-exp}

\input{tex/5-con}

\section{Acknowledgement}
This research is supported in part by National Key R\&D Program of China (2022ZD0115502), National Natural Science Foundation of China (NO. 62122010, U23B2010), Zhejiang Provincial Natural Science Foundation of China under Grant No. LDT23F02022F02, Key Research and Development Program of Zhejiang Province under Grant 2022C01082.

\clearpage
\bibliographystyle{IEEEtran}
\bibliography{IEEEexample}

\end{document}

%% file: tex/0-abstract.tex
\begin{abstract}
Recent 3D object detectors typically utilize multi-sensor data and unify multi-modal features in the shared bird’s-eye view (BEV) representation space. 
However, our empirical findings indicate that previous methods have limitations in generating fusion BEV features free from cross-modal conflicts. These conflicts encompass extrinsic conflicts caused by BEV feature construction and inherent conflicts stemming from heterogeneous sensor signals.
Therefore, we propose a novel Eliminating Conflicts Fusion (ECFusion) method to explicitly eliminate the extrinsic/inherent conflicts in BEV space and produce improved multi-modal BEV features. Specifically, we devise a Semantic-guided Flow-based Alignment (SFA) module to resolve extrinsic conflicts via unifying spatial distribution in BEV space before fusion. Moreover, we design a Dissolved Query Recovering (DQR) mechanism to remedy inherent conflicts by preserving objectness clues that are lost in the fusion BEV feature.
In general, our method maximizes the effective information utilization of each modality and leverages inter-modal complementarity. Our method achieves state-of-the-art performance in the highly competitive nuScenes 3D object detection dataset. The code is released at \url{https://github.com/fjhzhixi/ECFusion}.
\end{abstract}

%% file: tex/1-intro.tex
\section{introduction}

3D object detection is crucial for enabling safe and effective autonomous driving, allowing vehicles to locate and recognize objects in real-world 3D environments accurately. 
To achieve precise and dependable 3D object detection, some methods combine information from LiDAR point clouds and camera RGB images via various multi-modal fusion strategies. 
Specifically, point clouds provide accurate 3D localization information, while RGB images offer rich contextual details. Thus combining these complementary modalities improves the accuracy and robustness of 3D object detection. Recently, advanced methods~\cite{bevfusion_mit, bevfusion_pku} attempt to fuse LiDAR-Camera features in the unified bird's-eye view~(BEV) space.

\begin{figure}[t]
  \centering
  \includegraphics[width=1\linewidth]{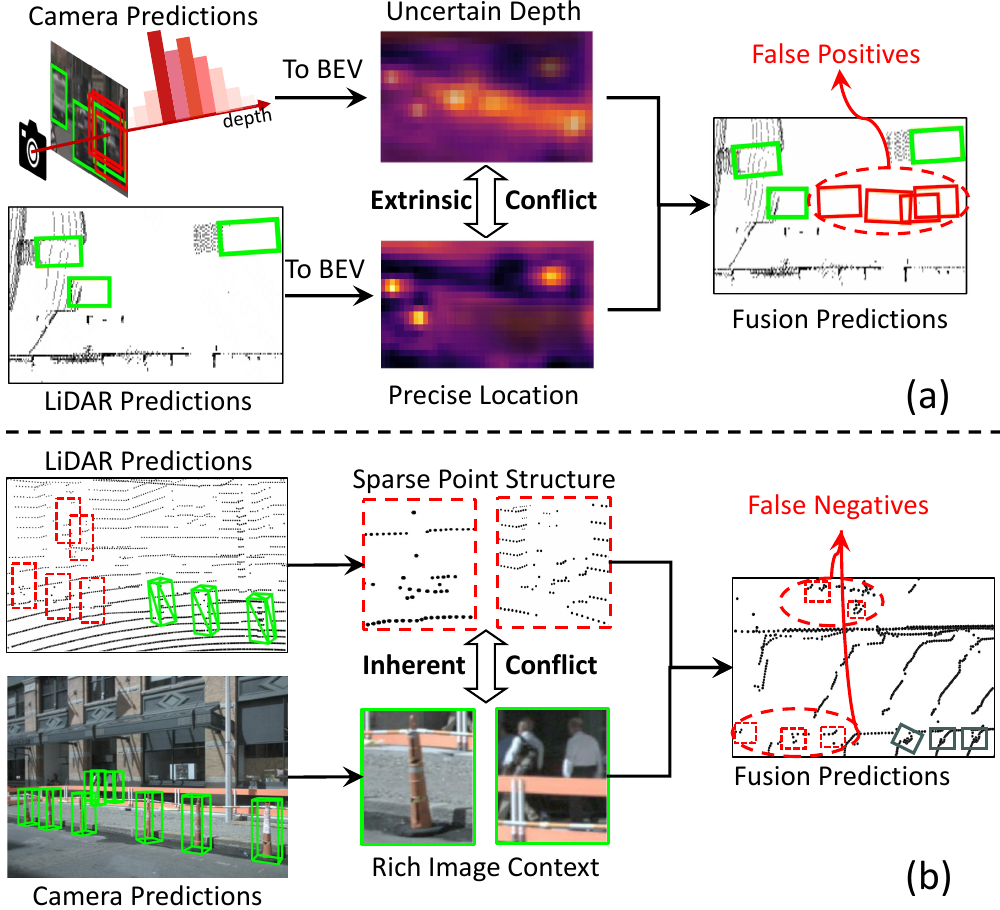}
  \vspace{-5mm}
  \caption{Cross-modal conflicts hinder LiDAR-Camera 3D object detection.
   \textcolor{green}{Green boxes} represents the correct prediction. \textcolor{red}{Red dotted boxes} and \textcolor{red}{red solid boxes}
   represent false negative and false positive respectively.
   (a) The single-modal LiDAR prediction is accurate, yet extrinsic conflicts caused by the uncertain depth of images lead to false positive results in fusion predictions. 
   (b) The single-modal camera prediction is accurate, yet inherent conflicts caused by the sparse points structure of small objects lead to false negative results in fusion predictions. 
   Best viewed in color.}
  \label{fig:motivation}
  \vspace{-8mm}
\end{figure}

In general, BEV space provides a suitable intermediate representation for multi-modal feature fusion. Nevertheless, existing fusion strategies only consider benefits brought by the complementary BEV features between modalities, while ignoring the interference caused by cross-modal conflicts. However, we argue that multi-modal fusion operations, influenced by conflicts among heterogeneous cross-modal features, potentially undermine accurate predictions.
In particular, the cross-modal conflicts are mainly from two aspects, \emph{i.e.}, \textbf{extrinsic conflict} and \textbf{inherent conflict}.
\textit{(\uppercase\expandafter{\romannumeral1})} The extrinsic conflicts arise from variations across different modalities in the construction process of BEV features. To elaborate, LiDAR and camera modalities exhibit spatially misaligned BEV feature distributions since they are extracted separately by independent encoders and mapped to BEV using different projection methods. Consequently, these misalignments inevitably lead to incorrect object information when merged. For instance shown in Fig.~\ref{fig:motivation}(a), 
the car that can be located correctly in LiDAR predictions has obviously misaligned spatial features in camera BEV since redundant objects are projected based on the uncertain depth. Such extrinsic conflicts from feature projection result in false positives in fusion predictions.
\textit{(\uppercase\expandafter{\romannumeral2})} The inherent conflicts stem from diverse patterns of sensor signals between modalities. Specifically, multi-modal features exhibit asymmetric perceptual capabilities for different objects, influenced by factors such as object distance, illumination, weather conditions, occlusion situation, \emph{etc}. Previous approaches expect that the modality with superior perceptual capabilities would dominate the fusion process. However, we reveal that excessively weak object confidence from the other modality can also hinder correct results. As shown in Fig.~\ref{fig:motivation}(b), 
the distant and small pedestrians and traffic cones can be recalled by the camera due to rich image visual clues but are missed in LiDAR predictions, which are limited by sparse point structure. Such inherent conflicts from sensor signals lead to false negatives in the fusion predictions.
Thence, cross-modal conflicts are non-negligible factors when utilizing multi-modal features to achieve accurate and robust detection.

In this paper, we propose an \textbf{E}liminating \textbf{C}onflicts \textbf{F}usion (\textbf{ECFusion}) method to avoid the degradation of perception capabilities caused by conflicts during fusion.
\textit{Firstly}, to eliminate extrinsic conflicts, we propose a \textbf{S}emantic-guided \textbf{F}low-based \textbf{A}lignment~(\textbf{SFA}) module, which aligns LiDAR and camera BEV features to the consistent distribution by utilizing spatial flow derived from semantic correspondence. Specifically, we first associate positions whose semantic information from class-aware heatmaps corresponds to the other modality. Then, the correspondence is converted into flow fields, which are employed to propagate BEV features for alignment. In this way, the fusion interference caused by extrinsic conflicts can be alleviated by alignment before fusion.
\textit{Secondly}, to eliminate inherent conflict, we introduce a \textbf{D}issolved \textbf{Q}uery \textbf{R}ecovering~(\textbf{DQR}) mechanism, which aims at discovering object queries that are dissolved in the fusion heatmaps due to inherent conflicts and recover them from individual LiDAR and camera BEV heatmaps. Concretely, besides generating object queries from the fusion heatmap as previous method~\cite{transfusion}, we also explore potential single-modal object queries. We focus on the positions that exhibit high objectness inconsistently with fused features via a masked heatmap strategy.
Our design is tailored to ensure maximum utilization of perceptual capability from single-modal features.
Our contributions are summarized as follows:
\begin{itemize}
    \item  We investigate ignored cross-modal conflicts when fusing multi-modal features to the unified BEV space and how they hinder LiDAR-Camera 3D object detection.
    \item We propose the ECFusion method to eliminate conflicts between multi-modal BEV features, including a SFA module for spatial alignment before fusion and a DQR mechanism to recover useful object queries after fusion.
    \item Extensive experiments show our method achieves state-of-the-art performance for LiDAR-Camera 3D object detection on the nuScenes dataset.
\end{itemize}

%% file: tex/2-relate.tex
\section{Related Work}
\subsection{Camera-based 3D Object Detection}
Camera-based 3D object detection aims to estimate the 3D bounding boxes of objects from 2D images. Methods~\cite{fcos3d, pgd, monopgc, gao2021room} focus on predicting 3D objects from monocular images by extending 2D detectors to 3D by regressing the depth and orientation of objects. Other methods~\cite{detr3d, petr,petrv2, wang2023object} adapt the query-based end-to-end paradigm of DETR~\cite{detr} to 3D space with 3D position queries and perspective projection.
Methods~\cite{bevformer, bevformerv2, polarformer, bevdet, bevdepth, m2bev} construct BEV feature from multi-view images via interaction with predefined grid-shaped BEV queries or view transformation~\cite{lss} to lift image features into voxel features.

\subsection{LiDAR-based 3D Object Detection}
LiDAR-based 3D object detection aims to localize 3D objects from point clouds scanned by LiDAR sensors.
Point-based methods~\cite{pointnet,pointnet++,pointrcnn,pv-rcnn,std,part-a2net,centerpoint} directly process point clouds without discretization or projection. They preserve fine-grained geometric information but face challenges such as irregularity and sparsity of point clouds, high computational complexity, and memory consumption.
Voxel-based methods~\cite{second,pointpillar,pixor,voxelnet, rangedet, wang2023costaware} convert point clouds into regular 3D/2D grids and encode to features by efficient 3D/2D convolution. They can reduce computational costs and memory consumption by sparsifying the voxel space but also introducing quantization errors and information loss.
Range view-based methods~\cite{rangercnn,rangedet} project 3D point clouds onto a spherical surface and encode attributes into pixel values. Range view is more compact and faster to process, but also includes some challenges such as scale variation, occlusion, and coordinate inconsistency.

\subsection{LiDAR-Camera Fusion-based 3D Object Detection}
Some methods~\cite{pointpainting, pointaugmenting, epnet, mvp, mmf, sfd, vpfnet, focalsparse, jacobson2023center} decorate point cloud features with the corresponding image information~(\emph{e.g.}, RGB values, context features, semantic categories) by mapping instructive clues from camera space to LiDAR space. Alternatively, some methods~\cite{emmf-det,Rpvne} project point clouds from LiDAR space to camera space and render images analogous to RGB-D data. 
However, these methods are mainly limited to the expression form of a particular modality and inevitably lose the information of another modality. 
Recently, some methods~\cite{bevfusion_mit, bevfusion_pku, transfusion,imvotenet, autoalign, autoalignv2, vff, gao2023sparse} attempt to fuse multi-modal features in the shared BEV space. Unlike the fusion methods above, they adopt two independent branches to encode inputs from the camera and LiDAR sensors into modality-specific features and unify multi-modal features into the BEV representation to fuse. Such a process preserves both the geometric structure from point clouds and semantic information from images. 

While existing BEV-based fusion methods have achieved remarkable strides, they still struggle with cross-modal conflicts in the BEV space. As a consequence, their performances are inferior even compared to single-modal detectors in some scenarios. To bridge these conflicts, we delve into exploring how specific conflicts disrupt the fusion process and pursue more fusion profit by eliminating these conflicts.

%% file: tex/3-method.tex
\section{methodology}
\begin{figure*}[ht]
  \centering
  \includegraphics[width=0.95 \linewidth]{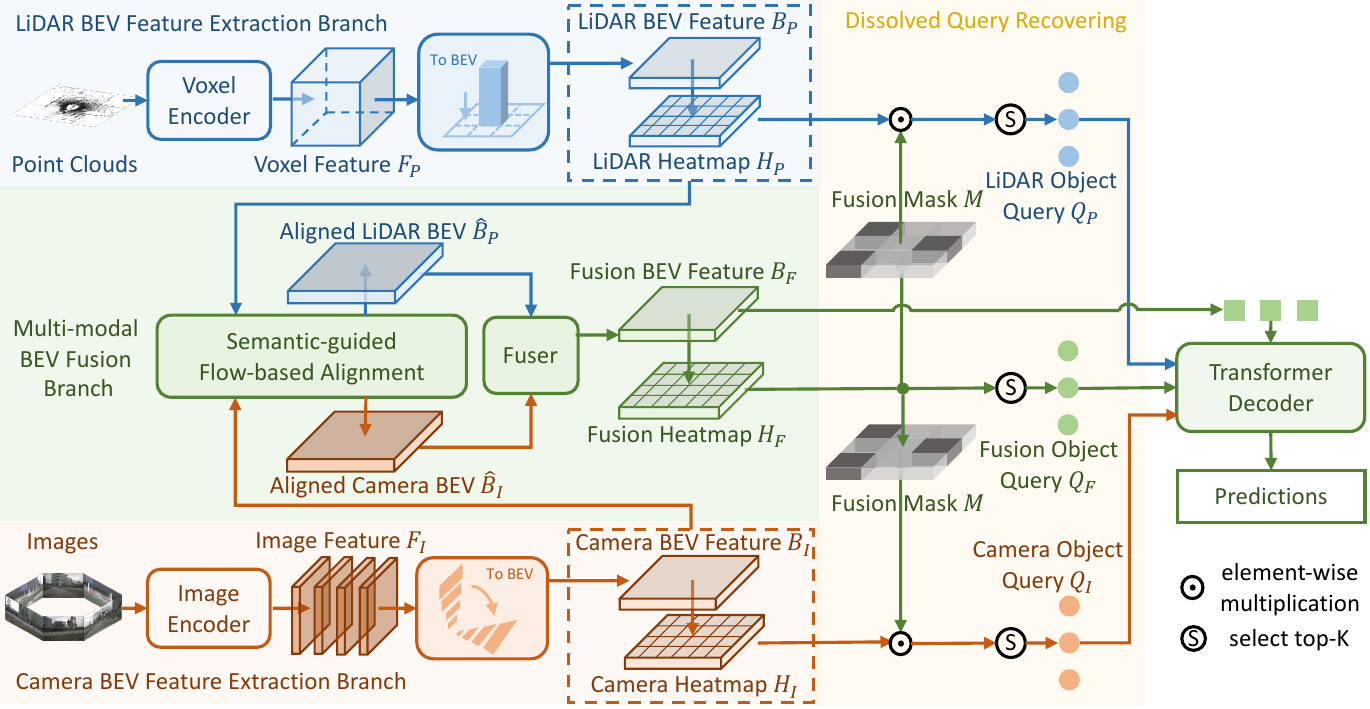}
  \vspace{-2mm}
  \caption{Overview of our framework. Given inputs of point clouds and multi-view images: (\uppercase\expandafter{\romannumeral1}) We process them by individual LiDAR/Camera Feature Extraction Branch to obtain modal-special BEV features $\mathbf{H}_P, \mathbf{H}_I,$. (\uppercase\expandafter{\romannumeral2}) We utilize Multi-modal BEV Fusion Branch to align $\mathbf{H}_P$ and $\mathbf{H}_I$ by SFA module and integrate them into unified fusion BEV features $\mathbf{H}_F$. 
  (\uppercase\expandafter{\romannumeral3}) We generate fusion object queries $\mathbf{Q}_F$ and modal-special object queries $\mathbf{Q}_P, \mathbf{Q}_I$ by DQR mechanism. 
  Finally, all queries are aggregated together to predict 3D bounding boxes through a transformer decoder.}
  \vspace{-3mm}
  \label{fig:method}
\end{figure*}

As shown in Fig.~\ref{fig:method}, our ECFusion method starts by utilizing LiDAR and camera BEV feature extraction branches to produce modal-specific BEV features from individual modalities.
Then, we leverage the multi-modal BEV feature fusion branch to integrate LiDAR and camera BEV features, forming a unified Fusion BEV feature. 
In the fusion branch, we propose a \textbf{S}emantic-guided \textbf{F}low-based \textbf{A}lignment~(\textbf{SFA}) module to first mitigate spatial distribution disparities (\emph{i.e.}, extrinsic conflict) between LiDAR and camera BEV features before fusing them.
Then, based on the LiDAR, camera, and fusion BEV features, we devise a \textbf{D}issolved \textbf{Q}uery \textbf{R}ecovering~(\textbf{DQR}) mechanism to generate comprehensive object queries. Concretely, the DQR mechanism aims to recover the dissolved object queries caused by asymmetric perceptual capabilities (\emph{i.e.}, inherent conflict) between two modalities from single-modal features.
Finally, a transformer decoder is utilized to predict the final 3D bounding boxes based on the derived object queries.

\subsection{Single-Modal BEV Feature Extraction}
\begin{figure}[t]
  \centering
  \includegraphics[width=0.9\linewidth]{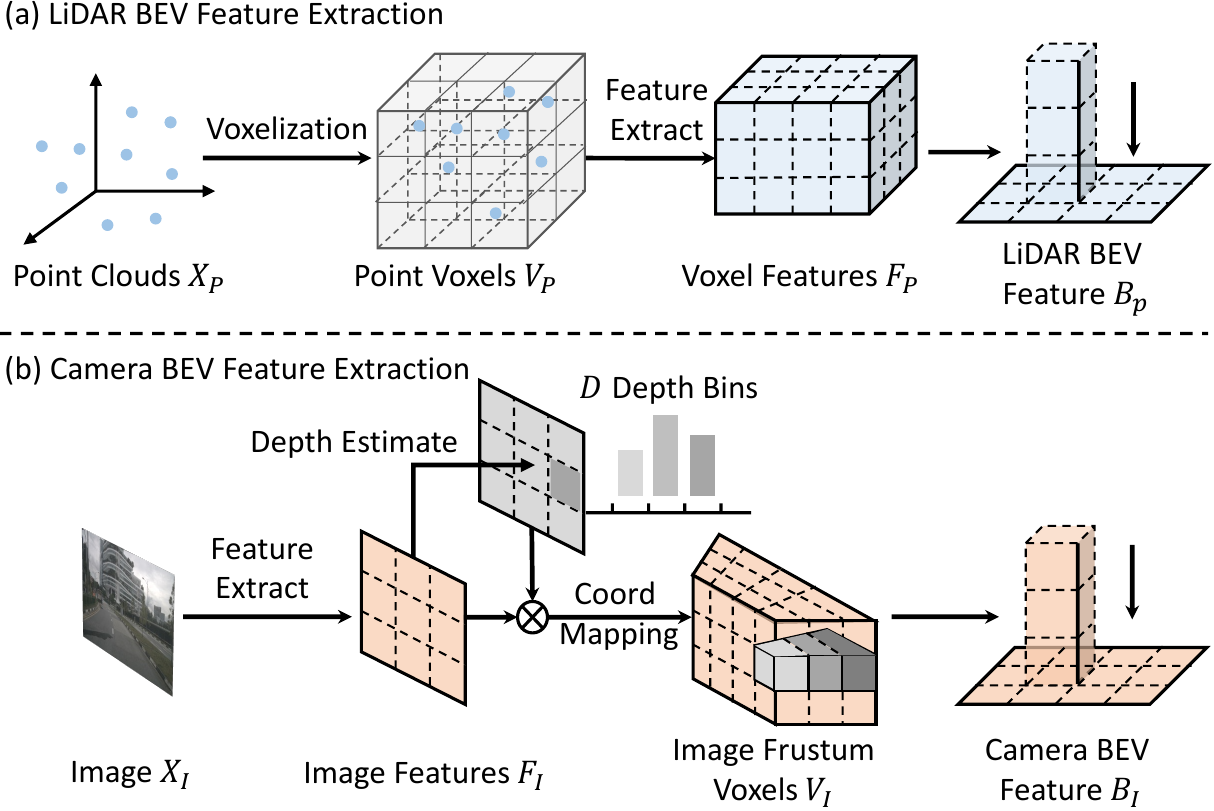}
  \vspace{-3mm}
  \caption{The LiDAR and camera BEV feature extraction process.}
  \vspace{-4mm}
  \label{fig:bevmap}
\end{figure}

The details of LiDAR and Camera BEV feature extraction branches are shown in Fig.~\ref{fig:bevmap}, in which LiDAR and Camera BEV features are produced separately. 

\textbf{LiDAR BEV Feature Extraction.} Given the input point clouds $\mathbf{X}_P$, as shown in Fig.~\ref{fig:bevmap}(a), we first partition them into regular voxels $\mathbf{V}_P \in \mathbb{R}^{X_v \times Y_v \times Z_v}$ and use voxel encoder with 3D sparse convolution~\cite{second} to extract features $\mathbf{F}_P \in \mathbb{R}^{X \times Y \times Z \times C}$, where $(X, Y, Z)$ denotes the size of the 3D voxel grids. Then we project $\mathbf{F}_P$ to BEV along the $Z$-axis and adopt several 2D convolutional Layers to obtain the LiDAR BEV feature map $\mathbf{B}_P \in \mathbb{R}^{X \times Y \times C}$. 

\textbf{Camera BEV Feature Extraction.} Given the input $N_v$ view images $\mathbf{X}_I$, as shown in Fig.~\ref{fig:bevmap}(b), we first extract features $\mathbf{F}_I \in \mathbb{R}^{N_v \times H \times W \times C}$ by image encoder, where $(H, W)$ denotes the size of the image feature map. To construct the camera BEV feature map $\mathbf{B}_I \in \mathbb{R}^{X \times Y \times C}$, we apply $2D\rightarrow3D$ view transform to image features of each view by Lift-Splat-Shoot~(LSS) module~\cite{lss}. 
Finally, The features within the same BEV grid are aggregated by BEVPool~\cite{bevfusion_mit} operation.

\subsection{Semantic-guided Flow-based Alignment}
In the multi-modal BEV feature fusion branch, we aim to combine the LiDAR BEV features $\mathbf{B}_P$ and camera BEV features $\mathbf{B}_I$ to construct fusion BEV features $\mathbf{B}_F$. 
However, previous methods~\cite{bevfusion_mit, bevfusion_pku} neglect the extrinsic conflicts between the two BEV features, \emph{i.e.}, inconsistent spatial/semantic distribution patterns, and directly concatenate the two BEV features as fusion results. Note that the extrinsic conflicts are caused by the difference in origin signals coordinate, BEV feature extraction pipeline, and projection process. 
For example, projecting image features to BEV space requires ill-posed monocular depth estimation, which inevitably predicts inaccurate object depth. As a result, with such an inaccurate object depth, the projected camera BEV feature will contain non-existent/redundant objects in the wrong positions, forming obvious misalignment with the LiDAR BEV feature.
%
\begin{figure}[t]
  \centering
  \includegraphics[width=0.9 \linewidth]{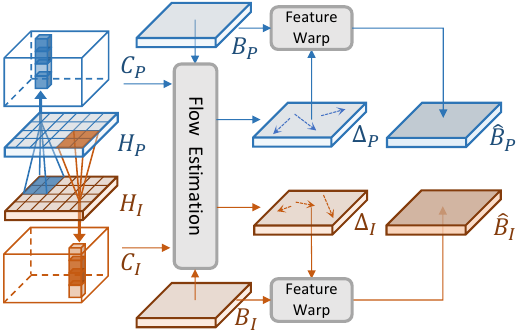}
  \caption{Semantic-guided flow-based alignment module.}
  \vspace{-4mm}
  \label{fig:alignment}
\end{figure}

Therefore, we elaborate a \textbf{S}emantic-guided \textbf{F}low-based \textbf{A}lignment~(\textbf{SFA}) module to align LiDAR and camera BEV features to obtain a consistent spatial distribution before fusion.
Inspired by the optical flow methods~\cite{deep_feature_flow, semtanic_flow}, we remedy the spatial discrepancy by applying appropriate flow transformation in inconsistent areas. 

Specifically, as shown in Fig.~\ref{fig:alignment}, we start by establishing spatial correspondence between two modalities. Since LiDAR and camera BEV features $\mathbf{B}_P,\mathbf{B}_I$ are generated from two individual/heterogeneous branches, it is impractical to directly establish the correspondence between $\mathbf{B}_P$ and $\mathbf{B}_I$. 
Thus we leverage the normalized LiDAR and camera BEV heatmaps $\mathbf{H}_{P}, \mathbf{H}_{I} \in \mathbb{R}^{X \times Y \times N_c}$, where $N_c$ is the number of object categories, to capture the pixel-wise spatial correspondence.
Technically, we construct the spatial correspondence for each pixel based on the cross-modal semantic similarity in the $q\times q$ neighborhood.

We first obtain cross-modal cost volumes $\mathbf{C}_{P}, \mathbf{C}_{I}  \in \mathbb{R}^{X \times Y \times q^2}$ from $\mathbf{H}_{P}, \mathbf{H}_{I}$, which can be formulated as:
\begin{small} 
\begin{equation}
\left\{\ \ 
\begin{aligned}
    & \mathbf{C}_P(i,j) = \mathop{\bigcup}\limits_{ {m \in [-q/2, q/2), \ n \in [-q/2, q/2)}} {\mathbf{H}_P(i,j)}^T \mathbf{H}_I(i+m, j+n), \\
    & \mathbf{C}_I(i,j) = \mathop{\bigcup}\limits_{ {m \in [-q/2, q/2), \ n \in [-q/2, q/2)}} {\mathbf{H}_I(i,j)}^T \mathbf{H}_P(i+m, j+n). \\
\end{aligned}
\right.
\end{equation}
\end{small}

Then we use a lightweight convolution block to estimate the flow fields $\Delta_{P}, \Delta_{I} \in \mathbb{R}^{X \times Y \times 2}$,  which serve as spatial correspondence between modalities:
\begin{equation}
    \{\Delta_{P}, \Delta_{I}\} = \mathrm{Conv}(\mathrm{Concat(\mathbf{B}_P, \mathbf{C}_P, \mathbf{B}_I, \mathbf{C}_I)}).
\end{equation}
Next, we adopt the differentiable bilinear sampling operation to warp features based on $\{\Delta_{P}, \Delta_{I}\}$, which linearly interpolates the features of neighbors around the warped position. Formally, the aligned BEV feature $\hat{\mathbf{B}}_{P}, \hat{\mathbf{B}}_{I}$ are obtained as:
\begin{equation}
    \hat{\mathbf{B}}_z(p) = \mathrm{Interp}(\mathbf{B}_z(p + \Delta_z(p)), \ \ \ z \in \{P, I\},
\end{equation}
where $\mathrm{Interp}(\cdot)$ denotes neighborhood bilinear interpolation.
Then we fuse them as $\mathbf{B}_F = \mathrm{Conv}(\mathrm{Concat}(\hat{\mathbf{B}}_{P}, \hat{\mathbf{B}}_{I}))$. Benefiting from the flow-based spatial alignment before fusion, it avoids uncoordinated features caused by extrinsic conflicts.

\subsection{Dissolved Query Recovering Mechanism}
Based on the fusion BEV feature $\mathbf{B}_F$, following~\cite{transfusion}, most previous methods directly generate class-specific fusion heatmap $\mathbf{H}_F \in \mathbb{R}^{X \times Y \times N_c}$ and select the Top-$K_F$ local maximum candidate indexes. The information of selected candidates is used to initialize the context feature and position embedding of object queries $\mathbf{Q}_F \in \mathbb{R}^{K_F \times C}$, which are used to aggregate relevant context and prediction box parameters through DETR-style decoder layers~\cite{detr}.

\begin{figure}[t]
  \centering
  \includegraphics[width=0.95\linewidth]{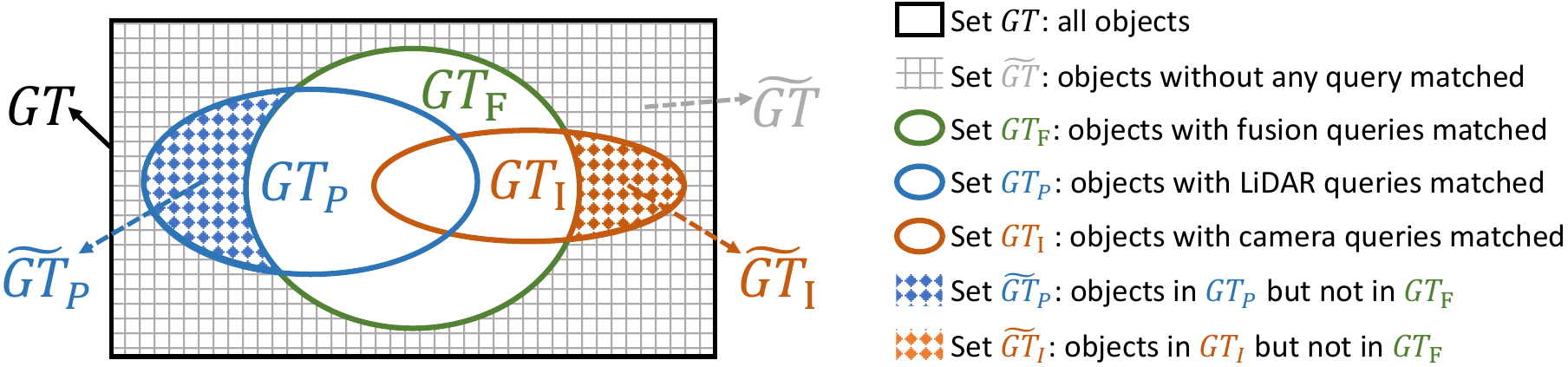}
  \vspace{-1mm}
  \caption{Analysis of queries set from LiDAR, camera, and fusion heatmaps. GT means ground-truth objects. Note that our DQR aims at recovering queries to match with $\widetilde{GT}_P$ and $\widetilde{GT}_I$.}
  \vspace{-5mm}
  \label{fig:query}
\end{figure}

Therefore, ensuring the high quality of initial queries is crucial to accurate detection since the object is unlikely to be recalled if there is no corresponding query. 
Theoretically, we expect fusion queries $\mathbf{Q}_F$ should integrate all valuable objectness clues from both point clouds and image modalities, enabling them to inherit the unique detection ability of each modality. 
However, we found that in current method $\{ {GT}_P \cup {GT}_I \} \not\subset {GT}_F$ as shown in Fig.~\ref{fig:query},
which means numerous objects in $\{ \widetilde{GT}_P \cup \widetilde{GT}_I \}$ that are not matched by fusion queries can be recalled by modal-special queries.
Our investigation shows that although the fusion strategy learned by current methods can indeed identify new objects, it would sacrifice a non-negligible portion of the single-modal detection ability. Therefore, it is pivotal to investigate how to maintain the single-modal detection capability while leveraging the cross-modal complementary. 

Hence, we propose a \textbf{D}issolved \textbf{Q}uery \textbf{R}ecovering~(\textbf{DQR}) mechanism to explicitly maintain the single-modal detection capability.
Our fundamental concept revolves around the exploration of queries that are dissolved in the fusion heatmap due to conflicts but can be recovered from single-modal heatmaps.
Further, we comprehensively integrate multi-source queries for high recall.

Concretely, In addition to queries $\mathbf{Q}_F$ from $\mathbf{H}_F$, we also generate modal-specific queries $\mathbf{Q}_P, \mathbf{Q}_I$ from single-modal heatmaps $\mathbf{H}_P, \mathbf{H}_I$ as complement.
We commence by generating a fusion mask $M$ based on the fusion query positions $\mathcal{P}_F$. By applying this mask to single-modal heatmaps, we circumvent redundant query generation for the same object, thereby focusing on the exploration of distinctive object positions $\mathcal{P}_P, \mathcal{P}_I$ compared to the fusion heatmap as below:
\begin{equation}
\left\{\ \ 
\begin{aligned}
    & \mathcal{P}_F = \mathrm{arg max}(\mathbf{H}_F, K_F), \\
    & \mathbf{M}[p] = \left\{
        \begin{aligned}
        & 0 \ \ \ if p \in \mathcal{P}_F, \\
        & 1 \ \ \ else,
        \end{aligned}
        \right. \\
    & \mathcal{P}_z = \mathrm{arg max}(\mathbf{H}_z \odot M, K_z) \ \ z \in \{P, I\}, \\
\end{aligned}
\right.
\end{equation}
where $\odot$ represents element-wise multiplication.
After obtaining multiple sets of query positions, we sample the corresponding BEV features with respective position embedding to construct the unified queries $\mathbf{Q}_u$ as: 
\begin{equation}
    \mathbf{Q}_u = \mathop{\bigcup}\limits_{z \in \{F, P, I\}} \{ \mathbf{Q}_z \}, \ \ \mathbf{Q}_z = \mathbf{B}_z[\mathcal{P}_z] + \mathrm{PE}(\mathcal{P}_z),
\end{equation}
where $\mathrm{PE}(\cdot)$ means position encoding function. Then in the transformer decoder, we adopt the self-attention between queries for relations between different objects and the cross-attention between queries and fusion features to aggregate relevant context. Finally, a feed-forward network~(FFN) is utilized for predicting 3D bounding boxes $\{ b_i\}_{i=1}^{K_F+K_P+K_I}$ according to the aggregated query features.

\subsection{Model optimization}
During training, the matching cost and loss function as \cite{bevfusion_mit} are applied. In detail, detection loss $L_{D}$ for all bounding box predictions is computed to optimize, including classification loss for all results and box regression loss for only positive pairs with Ground-Truth~(GT) boxes matched. 
Besides, we adopt focal losses $\{ L_{H_z} | z \in {F, P, I} \}$ for three heatmap predictions with the gaussian-distributed GT map generated by GT boxes center~\cite{centerpoint}. Noted that the GT map
for $L_P, L_I$ should be masked by fusion mask $M$ synchronously.
The total loss is $L = L_{det}+  \mathop{\sum}\limits_{z \in \{F, P, I\}} L_{H_z}$.

%% file: tex/4-exp.tex
\section{experiments}

\subsection{Datasets and Metrics}
We validate the effectiveness of our method on the large-scale nuScenes dataset\cite{nuscenes}, which is currently the most popular benchmark for multi-modal 3D object detection. 
nuScenes contains 700 sequences for training, 150 sequences for validation, and 150 sequences for testing. 
Each sequence is approximately 20 seconds long, including annotated 3D bounding boxes from 10 categories of sampled key frames. 
Each sample consists of point clouds from $32$-beam LiDAR scans and $6$ surround-view $1600 \times 900$ resolution images that provide $360^{\circ}$ horizontal FOV. 
For metrics, nuScenes provides the official evaluation metrics for the 3D detection task including mean Average Precision~(mAP) and nuScenes Detection Score~(NDS).
%

\subsection{Implementation Details}
We implement our single-modal BEV feature extraction branches following BEVFusoin~\cite{bevfusion_mit}. We set the image size to $448 \times 800$ as the official settings of the camera-based methods~\cite{bevdet, bevdepth} and the voxel size to $0.075$m as the official settings of the LiDAR-based methods~\cite{centerpoint, pointpillar}. Our training consists of two stages as BEVFusion: 1) We respectively train the LiDAR and camera branch with single-modal inputs. 2) We jointly train the entire model that loads weights from two pre-trained branches and frozen modal encoders. We set the query numbers to $K_F=200, K_P=50, K_I=50$ for training and testing. We directly use predictions from a single model without any test-time augmentation or extra post-processing for both val and test results.

\input{table/nuscene_result}
\input{table/ablations}

\subsection{Comparison with State-of-the-Art Methods}
We show the main results of nuScenes in Table~\ref{tab:nuscene_result}, our method surpasses all previous LiDAR-camera fusion methods and achieves the state-of-the-art performance of $73.4\%$ NDS on \textit{val} set and $73.9\%$ NDS on \textit{test} set. Our method has a significant improvement over LiDAR or camera methods. Moreover, By contrast with baseline method BEVFusion~\cite{bevfusion_mit}, our method exceeds it by a large margin with +$2.2\%$ mAP and +$2.0\%$ NDS on \textit{val} set. We ascribe this performance gain to eliminating cross-modal conflicts in BEV space.

\input{table/ablation_sfa}
\begin{figure}[t]
  \centering
  \includegraphics[width=0.9\linewidth]{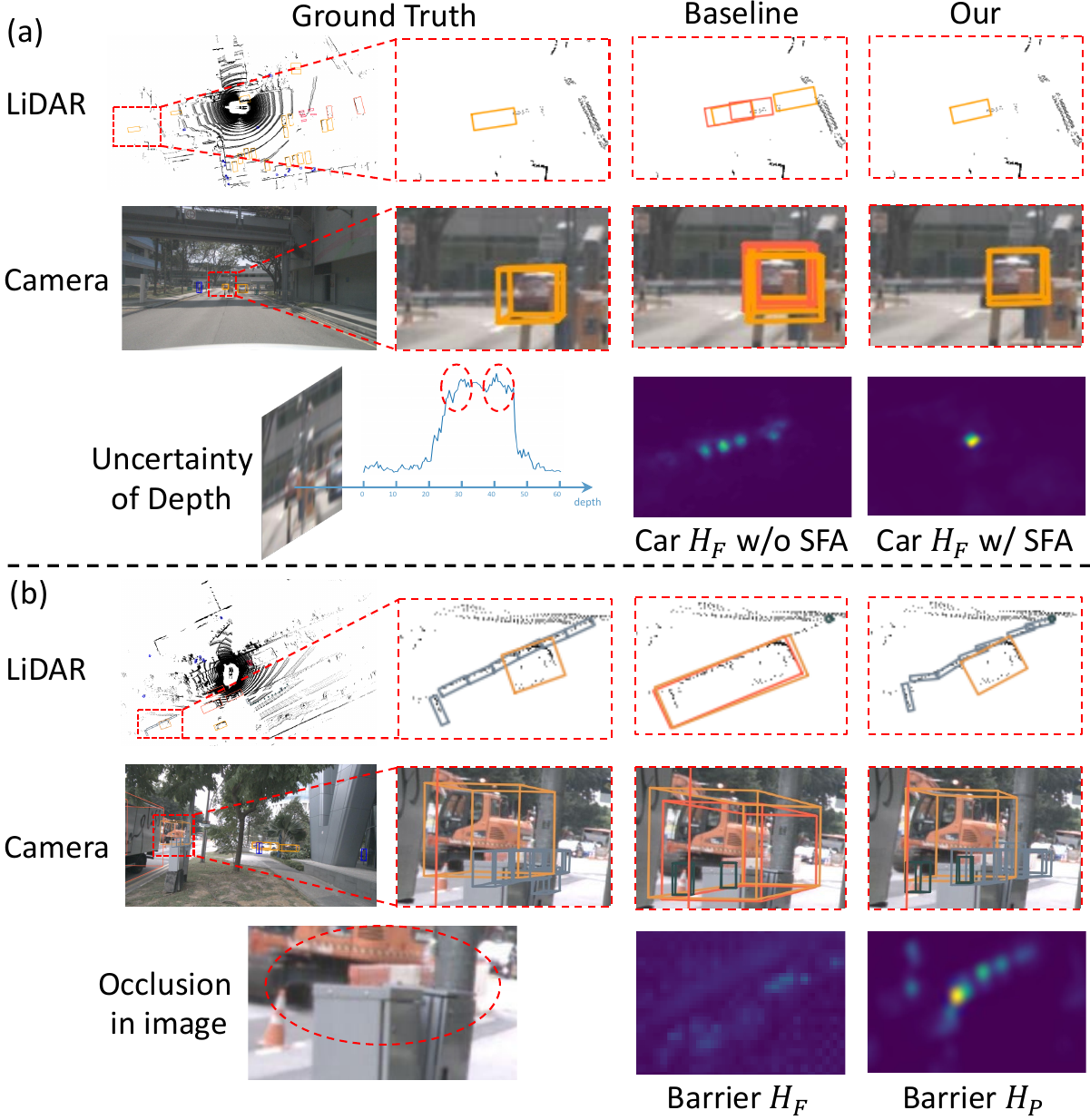}
  \vspace{-4mm}
  \caption{Visualization of comparing our method with baseline on nuScenes \textit{val} set. We show ground truth boxes on LiDAR and camera space respectively. For the baseline and our method, we show the predictions on LiDAR and camera space. Furthermore, we visualize the BEV heatmaps constructed by different methods.}
  \vspace{-4mm}
  \label{fig:exp_result}
\end{figure}

\subsection{Ablation Studies}
In this subsection, we present ablation studies and in-depth analyses of our designs. The mAP and NDS are evaluated on the nuScenes \textit{val} set.

\textbf{Multi Modalities}. We verify the effectiveness of multi-modal fusion in Table~\ref{tab:multi-modal} with improvements to LiDAR-based results~(+$6.1\%$ mAP and $+4.2\%$ NDS) and camera-based results~(+$34.6\%$ mAP and +$37.5\%$ NDS).

\textbf{Main Contributions}. To illustrate our improvements for fusion, we conduct ablation experiments on our two primary components SFA and DQR in Table~\ref{tab:main-design}.
As a result, our SFA module can improve the baseline by +$0.7\%$ mAP and +$0.5\%$ NDS.
Furthermore, the method with the DQR mechanism demonstrates significant improvements, elevating the baseline by +$2.1\%$ mAP and +$1.4\%$ NDS.
Meanwhile, the simultaneous utilization of our two primary design components synergistically enhances performance, resulting in a combined improvement of +$2.3\%$ mAP and +$1.7\%$ NDS.

\textbf{Qualitative Results}. To gain more insight into our improvements, we compare the qualitative results in Fig.~\ref{fig:exp_result}. In sub-fig. (a), Caused by the extrinsic conflicts from inevitable uncertainty of depth estimation for the car at a long distance, fusion BEV heatmap $\mathbf{H}_F$  has redundant high activations of the Car category. In contrast, our heatmap can obtain accurate object position assistance by the alignment of the SFA module.
Moreover, we show a case in sub-fig. (b) when the detection capability of the camera for these barriers is lost due to the occlusion, fusion BEV heatmap $\mathbf{H}_F$ of baseline loses valuable objectness of barriers due to inherent conflicts. Meanwhile, our method can recover object queries from LiDAR heatmap $\mathbf{H}_P$ to retain the right predictions.

\textbf{SFA Module}. To precisely demonstrate the superiority of our aligned modality-specific BEV features, we ablate the design of our SFA module as shown in Table~\ref{tab:abl-sfa}. Comparisons between \#2-\#4 and \#1 show that applying flow-based alignment to any modality can bring improvements, \emph{i.e.}, +$0.4\%$ NDS with LiDAR BEV flow $\Delta_P$ and +$0.7\%$ NDS with camera BEV flow $\Delta_I$. Both utilization can also further achieve significant +$1.1\%$ NDS improvements.
Moreover, we show in \#5 compared with \#4 that without semantic guidance represented by cost volume $\mathbf{C}_{P}, \mathbf{C}_{I}$ results in performance degradation -$0.6\%$ NDS.
Overall, our approach estimates multi-modal flows to facilitate cross-modal consistency that leads to improved fusion BEV features.

\input{table/ablation_query}
\textbf{DQR Mechanism}. We validate the efficacy of our DQR in Table~\ref{tab:abl-query}. 
Comparisons \#2-\#4 with \#1 reveal that results with recovering dissolved queries from single-modal heatmaps significantly outperform the baseline. For instance, LiDAR queries $\mathbf{Q}_P$ bring +$0.9\%$ NDS, and camera queries $\mathbf{Q}_I$ bring +$0.6\%$ NDS, while using both achieve an even higher gain of +$1.4\%$ NDS.
To confirm the necessity of fusion mask $\mathbf{M}$, we conduct experiment \#5, which independently generates three groups of queries from heatmaps without $\mathbf{M}$. The results as $-0.2\%$ NDS drop compared with \#1 highlight the necessity of our mask design.
Furthermore, simply increasing the number of queries from $200$ to $300$ in experiment \#6 yields only marginal improvements, indicating that our approach effectively discovers and recovers lost objects due to cross-modal conflicts, rather than using more queries.

%% file: table/nuscene_result.tex
\begin{table}
\caption{Comparison with state-of-the-art methods on the nuScenes {\tt validation} and {\tt test} set. We report metrics mAP(\%) and NDS(\%). Modality `{\tt L}' {and} `{\tt C}' represent LiDAR and camera. }
\vspace{-1mm}
\scriptsize
\renewcommand\tabcolsep{3.0pt}
\renewcommand\arraystretch{1.2}
\renewcommand{\footnote}{\fnsymbol{footnote}} 
\small
\setlength{\abovecaptionskip}{0.0cm}
\setlength{\belowcaptionskip}{-0.45cm}
\centering

\begin{tabular}{l|c|cc|cc}

\toprule

\multirow{2}*{Method} & \multirow{2}*{Modality} & \multicolumn{2}{c}{\textit{validation}}\vline & \multicolumn{2}{c}{\textit{test}} \\ 
& & mAP$\uparrow$ & NDS$\uparrow$ & mAP$\uparrow$ & NDS$\uparrow$ \\

\midrule

BEVDet4D~\cite{bevdet} & C & 42.1 & 54.5 & 45.1 & 56.9 \\

BEVFormer~\cite{bevformer} & C & - & - & 48.1 & 56.9\\

Ego3RT~\cite{ego3rt} & C & 47.8 & 53.4 & 42.5 & 47.9 \\

PolarFormer~\cite{polarformer} & C & 50.0 & 56.2 & 49.3 & 57.2 \\

\midrule

CenterPoint~\cite{centerpoint} & L & 59.6 & 66.8 & 60.3 & 67.3 \\

Focals Conv~\cite{focalsparse} & L & 61.2 & 68.1 & 63.8 & 70.0\\

Transfusion-L~\cite{transfusion} & L & 65.1 & 70.1 & 65.5 & 70.2 \\

LargeKernel3D~\cite{largekernel3D} & L & 63.3 & 69.1 & 65.3 & 70.5 \\

\midrule

FUTR3D~\cite{futr3d} & L+C & 64.5 & 68.3 & - & - \\

PointAug~\cite{pointaugmenting} & L+C & - & - & 66.8 & 71.0  \\

MVP~\cite{mvp} & L+C & 67.1 & 70.8 & 66.4 & 70.5\\

AutoAlign~\cite{autoalign} & L+C & 66.6 & 71.1 & - & - \\

AutoAlignV2~\cite{autoalignv2} & L+C & 67.1 & 71.2  & 68.4 & 72.4 \\

TransFusion~\cite{transfusion} & L+C & 67.5 & 71.3 &  68.9 & 71.6 \\

VFF~\cite{vff} & L+C & - & - &  68.4 & 72.4 \\

BEVFusion~\cite{bevfusion_pku} & L+C & 67.9 & 71.0 & 69.2 & 71.8 \\

UVTR~\cite{uvtr} & L+C & 65.4 & 70.2 & 67.1 & 71.1 \\

Focals Conv-F~\cite{focalsparse} & L+C & - & - & 67.8 & 71.8 \\

BEVFusion~\cite{bevfusion_mit} & L+C & 68.5 & 71.4  & 70.2 & 72.9 \\

DeepInteraction~\cite{deepinteraction}& L+C & 69.9 & 72.6 & 70.8 & 73.4 \\

\rowcolor{mygray} ECFusion(Our) & L+C & \textbf{70.7} & \textbf{73.4} & \textbf{71.5} & \textbf{73.9} \\

\bottomrule

\end{tabular}

\label{tab:nuscene_result}
\vspace{-1mm}
\end{table}

%% file: table/ablations.tex
\begin{table}[t]
\small
    \parbox{.47\linewidth}{
        \centering
        \setlength{\tabcolsep}{3pt}
        \caption{Ablation studies on the multi-modal.}
        \vspace{-1mm}
        \begin{tabular}{c|c|cc}
        \toprule
        LiDAR & Camera & mAP$\uparrow$ & NDS$\uparrow$ \\
        
        \midrule
        
        \checkmark & & 64.6 & 69.2\\
        & \checkmark & 36.1 & 35.9 \\
        \rowcolor{mygray} \checkmark & \checkmark & \textbf{70.7} & \textbf{73.4} \\
        
        \bottomrule
        \end{tabular}
        \label{tab:multi-modal}
    }
    \hfill
    \parbox{.47\linewidth}{
        \centering
        \setlength{\tabcolsep}{4pt}
        \caption{Ablation studies on main designs.}
        \vspace{-1mm}
        \begin{tabular}{c|c|cc}
        \toprule
        SFA & DQR & mAP$\uparrow$ & NDS$\uparrow$ \\
        
        \midrule
        
        & & 68.4 & 71.7   \\
        \checkmark & & 69.1 & 72.2 \\
        & \checkmark & 70.5 & 73.1 \\
        \rowcolor{mygray} \checkmark & \checkmark & \textbf{70.7} & \textbf{73.4} \\
        
        \bottomrule
        \end{tabular}
        \label{tab:main-design}
    }
    \vspace{-3mm}
\end{table}

%% file: table/ablation_sfa.tex
\begin{table}
\centering
        \caption{Ablation studies on the SFA module. {\tt $\Delta_P$} and {\tt $\Delta_I$} are flows for LiDAR and camera. {\tt $\mathbf{C}_{P}$} and {\tt $\mathbf{C}_{P}$} denote the cost volume.}
        \vspace{-1mm}
\begin{tabular}{c|cc|cc|cc}
\toprule
& $\Delta_P$ & $\Delta_I$ & $\mathbf{C}_{P}$ & $\mathbf{C}_{I}$ & mAP$\uparrow$ & NDS$\uparrow$ \\
\midrule
\#1 & & & & & 68.4 & 71.7 \\
\midrule

\#2 & \checkmark & & \checkmark & & 68.9 & 72.1\\
\#3 & & \checkmark & & \checkmark & 69.3 & 72.4 \\
\rowcolor{mygray} \#4 & \checkmark & \checkmark & \checkmark & \checkmark & \textbf{69.5} & \textbf{72.8} \\
\midrule

\#5 & \checkmark & \checkmark & & & 69.1 & 72.2 \\

\bottomrule

\end{tabular}
\label{tab:abl-sfa}

\end{table}

%% file: table/ablation_query.tex
\begin{table}
\centering
\caption{Ablation studies on the DQR mechanism. {\tt $\mathbf{Q}_P$} and {\tt $\mathbf{Q}_I$} are recovered queries from LiDAR and camera. {\tt $\mathbf{M}$} denotes our fusion mask. {\tt $K_F$} denotes the fusion query number.}
\vspace{-1mm}
\begin{tabular}{c|cc|c|c|cc}
\toprule & $\mathbf{Q}_P$ & $\mathbf{Q}_I$ & $\mathbf{M}$ & $K_F$ & mAP$\uparrow$ & NDS$\uparrow$ \\
\midrule

\#1 & & & & 200 & 68.4 & 71.7\\
\midrule

\#2 & \checkmark & & \checkmark & \multirow{3}*{$200$} & 70.1 & 72.6\\
\#3 & & \checkmark & \checkmark & & 69.6 & 72.3 \\
\rowcolor{mygray} \#4 & \checkmark & \checkmark & \checkmark & & \textbf{70.5} & \textbf{73.1}\\

\midrule
\#5 & \checkmark & \checkmark & & $200$ & 68.0 & 71.5 \\
\midrule
\#6 & & & & $300$ & 68.7 & 71.8 \\

\bottomrule
\end{tabular}
\vspace{-3mm}
\label{tab:abl-query}

\end{table}

%% file: tex/5-con.tex
\section{Conclusion}
In this work, we present a novel approach to eliminate widely prevalent yet long-overlooked cross-modal conflicts during the fusion process for LiDAR-Camera 3D object detection. This key idea is to align various spatial distributions before fusion and recover lost object information after fusion. To achieve that, we introduce an SFA module to exploit cross-modal correspondence and align multi-modal features via a spatial flow field. Moreover, we design a DQR mechanism to persevere object queries that are obscured in fusion features due to conflict interference. Experiments on the nuScenes dataset prove the effectiveness of our method.